\pdfoutput=1

\documentclass[11pt]{article}

\usepackage[preprint]{acl}

\usepackage{times}
\usepackage{latexsym}
\usepackage{algorithm}
\usepackage{algpseudocode}
\usepackage[T1]{fontenc}

\usepackage[utf8]{inputenc}

\usepackage{microtype}

\usepackage{amsmath,amsfonts,bm}

\def\eqref#1{equation~\ref{#1}}

\def\1{\bm{1}}

\DeclareMathAlphabet{\mathsfit}{\encodingdefault}{\sfdefault}{m}{sl}
\SetMathAlphabet{\mathsfit}{bold}{\encodingdefault}{\sfdefault}{bx}{n}

\DeclareMathOperator*{\argmax}{arg\,max}

\usepackage{hyperref}
\usepackage{url}
\usepackage{booktabs} %
\usepackage{graphicx}
\usepackage{multirow}
\usepackage{xcolor}
\usepackage{float}
\usepackage{wrapfig}
\usepackage{bbm}
\usepackage{subcaption}

\usepackage[noabbrev,capitalize]{cleveref} %

\crefname{equation}{equation}{equations}   %
\crefname{footnote}{footnote}{footnotes}   %
\crefname{line}{line}{lines}               %
\crefname{section}{\S}{\S\S}
\Crefname{section}{\S}{\S\S}    %
\crefformat{section}{#2\S#1#3}  %
\Crefformat{section}{#2\S#1#3}
\crefrangeformat{section}{\S\S#3#1#4--#5#2#6}
\Crefrangeformat{section}{\S\S#3#1#4--#5#2#6}

\title{Few-Shot Recalibration of Language Models }

\author{Xiang Lisa Li \\
  Stanford University \\
  \texttt{xlisali@stanford.edu} \\\And
  Urvashi Khandelwal \\
  Google DeepMind \\
  \texttt{urvashik@google.com} \\\And
  Kelvin Guu \\
  Google DeepMind \\
  \texttt{kguu@google.com} \\}

\newcommand{\exper}[1]{\textsc{#1}}
\newcommand{\plm}{p_\exper{LM}}
\newcommand{\ECE}{\exper{ECE}}

\newcommand{\Slice}{\exper{Slice}} 
 
\newcommand{\acc}{\text{acc}}
\newcommand{\conf}{\text{conf}}
\newcommand{\calibrator}{g_\theta}
\newcommand{\precision}{\text{prec}}
\newcommand{\countt}{\text{count}}

\newcommand{\setB}{\mathcal{B}}

\newcommand{\OursShort}{\textsc{FSC}}

\newcommand{\SampleAvg}{\text{Sample Avg}}

\newcommand{\DomainAvg}{\text{Domain Avg}}
\newcommand{\OW}[1]{\textcolor{gray}{#1}}

\begin{document}

\maketitle

\begin{abstract}

Recent work has uncovered promising ways to extract \emph{well-calibrated} confidence estimates from language models (LMs), where the model's confidence score reflects how likely it is to be correct. 
However, while LMs may appear well-calibrated over broad distributions, this often hides significant miscalibration within narrower slices (e.g., systemic \emph{over}-confidence in math can balance out systemic \emph{under}-confidence in history, yielding perfect calibration in aggregate).

To attain well-calibrated confidence estimates for any slice of a distribution, we propose a new framework for \textbf{few-shot slice-specific recalibration}. 
Specifically, we train a recalibration model that takes in a few \emph{unlabeled} examples from any given slice and predicts a curve that remaps confidence scores to be more accurate for that slice. 
Our trained model can recalibrate for arbitrary new slices, \emph{without using any labeled data from that slice}. This enables us to identify domain-specific confidence thresholds above which the LM’s predictions can be trusted, and below which it should abstain.  
Experiments show that our few-shot recalibrator consistently outperforms existing calibration methods, for instance improving calibration error for PaLM2-Large on MMLU by 16\%, as compared to temperature scaling.\footnote{Code is available at \url{https://github.com/XiangLi1999/FewShotCalibration.git}}\looseness=-1

\end{abstract}

\section{Introduction} 

Knowing when to trust a model’s predictions is typically mapped to the concept of calibration, where the model’s confidence estimate for a prediction reflects how likely it is to be correct.
Language models (LMs) have recently been shown to be well-calibrated in a number of settings \citep{kadavath2022language,xiao2022uncertainty,kuhn2023semantic,openai2023gpt4}.
However, while models can be well-calibrated for aggregate distributions (e.g. mixtures of a number of domains), they can be significantly miscalibrated for narrower domains within that distribution \citep{Yu2022RobustCW, pmlr-v80-hebert-johnson18a}.

\begin{figure*}
    \centering
    \includegraphics[width=\textwidth, page=1]{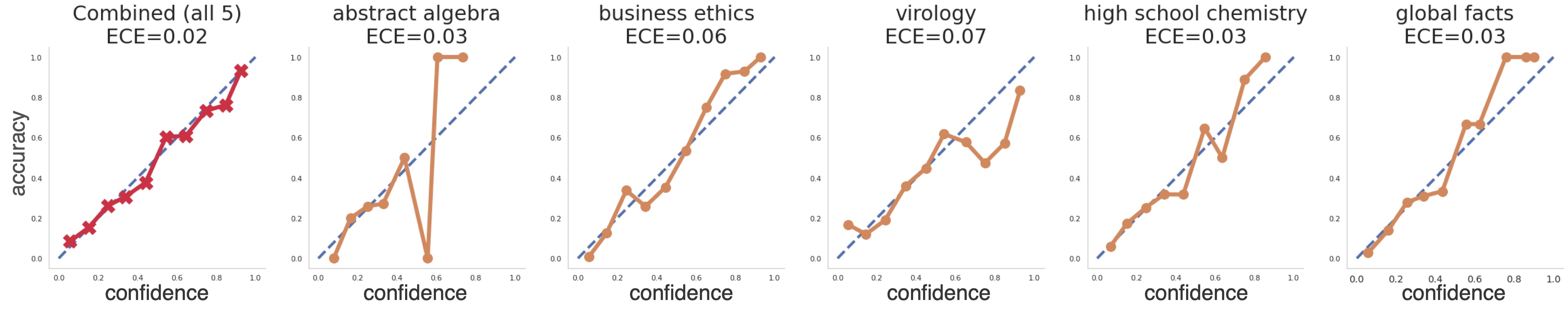}
    \vspace{-0.6cm}
    \caption[1]{\label{fig:observation} An example of the illusion of LM calibration. For a combination of five domains, the model is well-calibrated with a calibration error of 0.02 (the first plot). However, the same model is miscalibrated on the the five individual domains, each with a higher calibration error. \footnotemark{}}
    \vspace{-0.5cm}
\end{figure*}

For instance, Figure~\ref{fig:observation} shows an LM that is well-calibrated on the combined distribution of five domains, achieving near perfect calibration curve with low expected calibration error (ECE).  
However, curves for the individual domains appear significantly miscalibrated in comparison, with the least calibrated domain \emph{virology} having a 250\% higher calibration error. 
This miscalibration problem is hidden for the combined distribution because overconfidence in some domains cancels out underconfidence in others.

This illustrates a key problem: LMs are not well-calibrated for meaningful slices of broader distributions.
This is particularly relevant in practice where users querying an LM rarely sample from a broad combination of distributions at any given time, and are more likely to sample from slices like \emph{abstract algebra} or \emph{virology}.
Our goal is to recalibrate LMs for each of these fine-grained slices of a distribution, thereby allowing users to reliably determine when predictions can be trusted. \looseness=-1

\footnotetext{Although a smaller sample size in MMLU can cause some jaggedness, our experiments on XNLI confirm this finding for larger sample sizes as well.}

To recalibrate a model in these finer-grained settings, we propose \emph{slice-specific} \emph{few-shot} recalibration---a  new framework that uses only a small number of \emph{unlabeled} examples from a given slice to recalibrate the LM for that slice. 
Specifically, for a given LM, we train a separate recalibration model that takes a few unlabeled examples as input and outputs a curve that maps the LM's confidence scores to slice-specific estimates of precision (i.e. the percentage of examples above the given confidence score that will be correct). 
This precision curve can be used to achieve many downstream goals. For instance, we can identify the confidence threshold that achieves a minimum level of precision, to control the LM's error rate for this slice.
We can also transform the precision curve into the corresponding calibration curve and reduce calibration error on this slice (\cref{ssec:pc}).

To train our few-shot recalibration model, we employ a synthetic data generation strategy: given a corpus of \emph{labeled examples} that have been partitioned into different domains, we can synthetically construct many different slices by taking different weighted mixtures of these domains. For example, 80\% \emph{abstract algebra} and 20\% \emph{virology} from MMLU (\cref{ssec:synthetic_data}). 
Since we're working with labeled data, we can directly compute an LM's ground-truth precision curve on that slice.
Then, we train our recalibration model to predict this ground-truth precision curve, when only given access to a  random sample of \emph{unlabeled} examples from that slice (\cref{ssec:model}).
At inference time, our trained recalibrator generalizes to previously unseen slices, and only requires unlabeled data. \looseness=-1

We train our slice-specific calibrator to recalibrate LLaMA-65B \citep{touvron2023llama} and PaLM2-Large \citep{anil2023palm} on the MMLU \citep{hendrycks2021measuring} and XNLI \citep{conneau2018xnli} datasets, which already categorize examples into domains, allowing us to easily create slices. 
We evaluate our few-shot recalibrator against a variety of baselines in two settings: (1) achieving a desired level of target precision by identifying slice-specific confidence thresholds and (2) reducing calibration error per slice. 
Overall, we find that our slice-specific recalibrator consistently outperforms existing methods for calibration in all settings, and it extrapolates well to domains that are unseen at training time. 
For PaLM2-Large on MMLU, our calibrator achieves a 21\% higher success rate for achieving a target precision of 90 and a 16\% lower calibration error on the test set slices, compared to the standard method of using the precision/calibration curve of the \emph{combined distribution} over all domains.

\vspace{-0.15cm}
\section{The Illusion of LM Calibration}
\label{sec:issue}
Calibration is a key tool for knowing when language model predictions can be trusted and when they should abstain or defer to experts. 
However, calibration on an individual domain can be much worse than the aggregate data distribution \cite{Yu2022RobustCW, pmlr-v80-hebert-johnson18a}.  In this paper, we show that large language models suffer from the same calibration failure. While LMs appear to be well-calibrated on average, they are significantly miscalibrated in finer-grained settings.

We study LM calibration for multiclass classification: 
let $x$ be an input query drawn from some query distribution $p(x)$ and $y \in \{1, \cdots, K\}$ be the output class. 
Let $\plm(y \mid x)$ denote the model probability, which is also the model's confidence. 
Let $\hat{y} = \argmax_y \plm(y \mid x)$ be the model's prediction, and $y^*$ be the ground truth label. 

\subsection{Measuring Calibration}
\label{ssec:ece}
Calibration expresses how closely a model's confidence estimate for a prediction is aligned with the true probability that the prediction is correct, as measured by accuracy.
We use $\acc(\setB) = \mathbb{E}_{(x, y^*, \hat{y}) \in \setB} \mathbbm{1}(\hat{y} = y^*)$ to denote the model's accuracy for the set $\setB$, and $\conf(\setB) =\mathbb{E}_{(x, y^*, \hat{y}) \in \setB} \plm(\hat{y} \mid x) $ denotes the model's confidence on this set. 

\paragraph{Expected Calibration Error (ECE)}
This is the canonical metric which measures $L_1$ distance between the confidence and accuracy \citep{naeini2015obtaining}. To measure ECE, we first group all the $N$ predictions into $M$ equally sized bins based on their confidence estimates, denoted as $B_1 \cdots B_M$. We then calculate the average confidence and accuracy of each bin, and compute the ECE of the LM under this query distribution $p(x)$:
\vspace{-0.2cm}
\begin{align*}
    \ECE (\plm, p) = \sum_{i=1}^M \frac{|B_i|}{N}|\conf(B_i) - \acc(B_i) |
\end{align*}
\vspace{-0.4cm}

Perfectly calibrated models have $\ECE = 0$ i.e. model confidence matches expected accuracy at all confidence levels.
For example, suppose there are 100 examples, each with confidence 0.8. We expect that 80 of the examples are correctly classified. 

\vspace{-0.15cm}
\paragraph{Calibration Curves}
Also known as reliability diagrams, these curves are a visual representation of model calibration, plotting the expected model accuracy as a function of model confidence \citep{degroot1983forecasters,niculescu2005predicting}. 
Well-calibrated models lie close to the diagonal ($y=x$). 
\cref{fig:observation} shows example curves with respect to different query distributions $p(x)$.

\subsection{Miscalibration on Slices of Distributions}
\label{ssec:observation}
\begin{figure}
  \centering
  \vspace{-0.5cm}
  \includegraphics[width=0.7\linewidth]{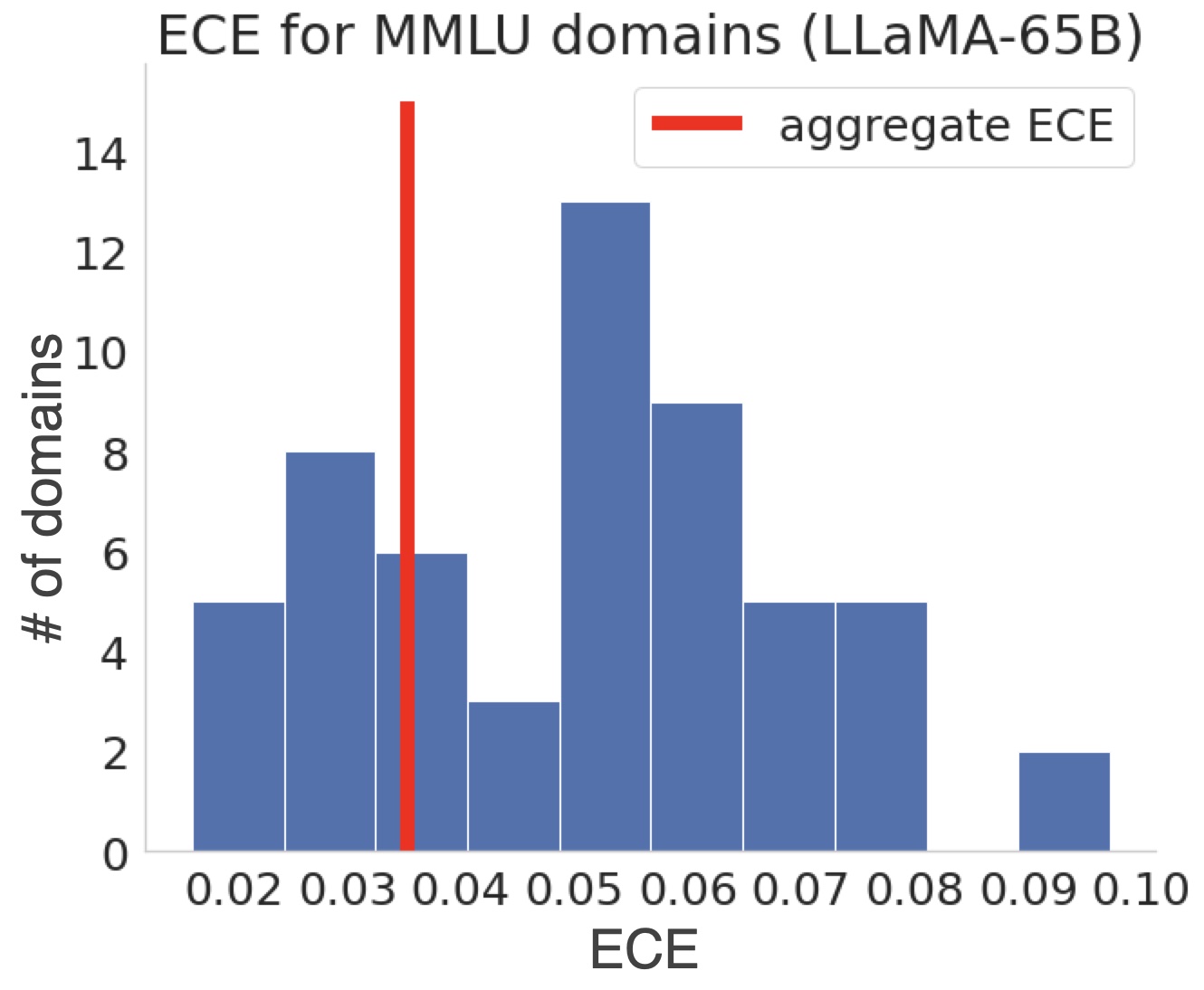}%
    \vspace{-0.35cm}
  \caption{\label{fig:quantative_observation} A histogram of ECE scores for LLaMA-65B on 57 MMLU domains. The red line shows ECE for all the domains combined. 
We can see the aggregate ECE is lower than most domains, hiding the underlying miscalibration problem.
    }
  \vspace{-0.35cm}
\end{figure}

Researchers often study LM calibration for aggregate query distributions ($p$), which are often composed of mixtures of meaningful finer-grained distributions: $p(x) = \sum_{d \in \mathcal{D}} \alpha_d p_d (x)$, where $\mathcal{D}$ denotes a set of domains, and each $p_d$ denotes the distribution of domain $d$, with relative frequency $\alpha_d$. 
For instance, \citet{openai2023gpt4} and \citet{kadavath2022language} have reported LM calibration on MMLU, which consists of 57 individual domains like \emph{abstract algebra}, \emph{high school chemistry} etc.
However, in practice, users querying an LM at a given point rarely sample from a broad aggregate distribution. They are more likely to sample from meaningful slices, like queries from \emph{abstract algebra} alone. \citet{Yu2022RobustCW, pmlr-v80-hebert-johnson18a} have shown that individual domains often suffer from miscalibration problem even if the aggregate distribution appears well-calibrated.

To demonstrate the same phenomenon for language models, we measure calibration of LLaMA-65B on combined MMLU ($p$), and also on each domain separately. As expected, the model is well-calibrated on $p$.
However, the LM is significantly miscalibrated for most domains.
This is shown in (\cref{fig:quantative_observation}) where the aggregate ECE is lower than that of most domains. 
It appears that the miscalibration problem is hidden for the broader distribution because overconfidence in some domains cancels out underconfidence in others.
\Cref{fig:observation} shows a qualitative example to illustrate the same miscalibration issue. 
These results show that LMs are not well-calibrated for meaningful slices of a broad distribution, leading us to address the problem via few-shot, slice-specific recalibration.

\vspace{-0.15cm}
\section{Slice-Specific Few-Shot Recalibration}
\vspace{-0.15cm}

Since individual fine-grained slices may be miscalibrated, we propose to recalibrate each slice. Intuitively, given a few samples from a slice, we can infer the rough identity of that slice, and then appropriately adjust the LM's confidences based on the LM's familiarity with the slice. For example, in practice, the first few queries in a user’s session can provide a sketch of the user’s query distribution (e.g., questions about abstract algebra). \looseness=-1

We formalize the task of slice-specific recalibration as learning a few-shot recalibrator $\calibrator\colon x_{1:k} \xrightarrow[]{} f$, which takes as input few-shot unlabeled samples $x_1 \cdots x_k$ drawn from a slice $p_i(x)$ and outputs a function $f$ that maps from raw confidence to adjusted confidence for this query distribution $p_i(x)$. The goal is for the recalibrator  $\calibrator$ to minimize the expected calibration error under different slices $p_i(x)$ after recalibration with $f$. Note that $f$ does not change the prediction of the underlying model $\plm$, only its confidences.

Next, we will discuss our algorithm for learning $\calibrator$. We discuss  our parametrization for output $f$ (\cref{ssec:pc}), how to synthesize training data that covers diverse slices (\cref{ssec:synthetic_data}), and how to train our recalibrator $\calibrator$ on this data (\cref{ssec:model}).

\subsection{Parametrizing $f$: Predicting Precision Curves v.s. Calibration Curves} 
\label{ssec:pc}
Recall that $f = \calibrator(x_{1}\cdots x_k)$ is the prediction target of our recalibrator, which will guide the adjustment of model's raw confidence.  
The most direct choice for $f$ would be the calibration curve (also known as the reliability diagram), i.e. a function that adjusts model confidence to predicted accuracy.
However, as described in \cref{ssec:ece}, calibration curves rely on binning predictions based on confidence estimates. This binning step introduces two hyperparameters: (1) the binning design, where scores can either be grouped into equally-spaced bins with equal interval ranges, or equally-sized bins with an equal number of examples per bin. And, (2) the number of bins such that scores can be grouped into a large number of bins each containing a small number of examples, or a small number of bins each containing many examples.
Both hyperparameters affect the shape of the calibration curve, and certain choices can hide miscalibration issues, making this an unreliable prediction target for the recalibrator.

Instead, we follow the practice of \citet{gupta2021calibration} and define $f$ to be the precision curve (PC; $\precision(\cdot)$), which maps confidence thresholds to precision scores. So, $\precision(0.5)=0.8$ means that for all the examples with confidence greater than 0.5, the model $\plm$ achieves a precision of 0.8.
In contrast to the calibration curve, the precision curve has no hyperparameters.
It is also extremely flexible. For instance, it can be converted to the corresponding calibration curve with any hyperparameter setting, given additional information about the distribution over confidence scores (see details in \cref{ssec:eval}). Conversely, it is hard to convert a calibration curve to a precision curve since the binning step is lossy.
This flexibility allows us to accomplish a variety of downstream goals such as reducing calibration error, finding optimal confidence thresholds for desired precision etc. as described in \cref{ssec:eval}.
For this reason, we choose precision curves as our calibrator's prediction target $f$. \looseness=-1

\vspace{-0.15cm}
\subsection{Synthetic Data Construction} 
\label{ssec:synthetic_data}
We now detail how we construct $(x_1 \cdots x_k, f)$ pairs to train our recalibrator. 
Each training example corresponds to a slice that must be recalibrated, and we must construct diverse slices to generalize to new slices at test time. We construct such slices by forming different mixtures of different domains (e.g. 80\% biology + 20\% history). 
This training data construction strategy scales beyond the number of domains by introducing more degrees of freedom: the number of mixture components, the choice of mixture, and the mixture weights.

\Cref{alg:1} shows how to construct one training example. To construct one slice, we first decide the number of domains in the slice by sampling an integer $m$ from a geometric distribution, then we randomly select those $m$ domains from the full set, and sample their mixture weights from a Dirichlet distribution. 
Once we have constructed the slice, we sample $k$ unlabeled examples from it to serve as the $x_{1:k}$ few-shot examples that provide a sketch of the corresponding slice. Then, we compute the ground-truth precision curve $f$ for this slice by using the model's predictions and the ground-truth labels for each example (\cref{ssec:pc}).  \looseness=-1

\begin{algorithm}
\caption{Synthetic Data Construction}\label{alg:1}
\begin{algorithmic}
\State \text{Sample $m \sim \text{Geo}(0.2)$ domains:}  $p_1 \cdots p_m$
\State \text{Sample mixure weights} $\alpha \sim \text{Dir}(1)$

\State \text{Sample examples}
\State  ~~~~~~ $\{(x_n, y_n)\}_{n=1}^N \sim  \Slice=\sum_{i=1}^m \alpha_i p_i$
\State \text{Predict} 
$\hat{y}_n = \plm(x_n)$ \text{for each $n=1\cdots N$}
\State \text{Compute precision curve $f$ from $\{x_n, y_n, \hat{y}_n\}_{n=1}^N$}
\State \text{Set $x_1 \cdots x_k$ as few-shot unlabeled samples}
\State \Return $(x_1 \cdots x_k), f$

\end{algorithmic}
\end{algorithm}
\vspace{-0.2cm}

\begin{figure*}
    \centering
    \includegraphics[width=1.0\textwidth, page=1]{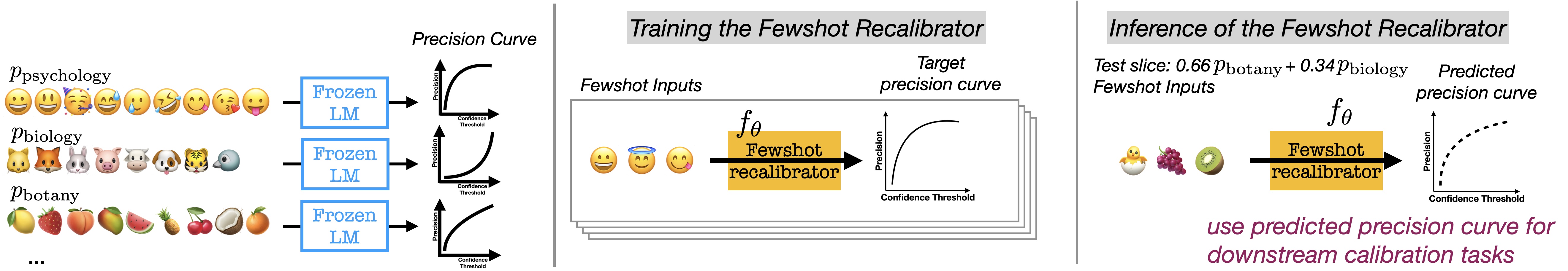}  

    \caption{\label{fig:method} 
    An illustration of the few-shot recalibrator. This model learns to predict the precision curve for slices (e.g. psychology only, or 20\% psychology-80\% biology) of a broader distribution (mix of psychology, biology, botany etc.), using few-shot unlabeled examples. At test time, it can predict the precision curve for an unseen slice (e.g. 66\% botany-34\% biology) given only an unlabeled few-shot set drawn from it. This precision curve can then be used to accomplish various downstream goals.
    }
\vspace{-0.4cm}
\end{figure*}

\subsection{Training the Few-Shot Recalibrator}
\label{ssec:model} 
Recall that we train our few-shot recalibrator $\calibrator$ that takes $k$ unlabeled examples $(x_1 \cdots x_k)$ and predicts the precision curve $f$ of the constructed slice. Concretely, we approximate the precision curve $f$ by predicting the precision score at 10 evenly spaced confidence thresholds: $[f(0.1),f(0.2), \cdots f(1.0)]$, and then linearly interpolate between these predicted values.  The training loss minimizes $L_2$ distance between the ground-truth and predicted precision at these 10 thresholds.

While the training loss penalizes all errors equally, over-estimating precision at some confidence threshold can be seen as a more costly error than under-estimating it.
This is because predicting a higher precision score than the ground-truth means the recalibrator believes the model correctly answers more questions than it actually can, and the confidence threshold does not trigger abstention when it should.
Conversely, when under-estimating precision, the confidence threshold is more conservative and sacrifices recall in favor of more reliable answers.
In this work, we prioritize correctness over recall, as is likely in most practical scenarios, 
by adapting the $L_2$ objective to be asymmetric: 
\begin{align*}
\small
\vspace{-0.2cm}
   & \mathcal{L}(\theta, c) = 
    \begin{cases}
\beta ||\hat{f}(c) -  f(c) ||^2 & \text{if } \hat{f}(c) >  f(c), \\
~~~||\hat{f}(c) -  f(c) ||^2 & \text{otherwise}.
\end{cases}\\
& \mathcal{L}(\theta) = \mathbb{E}_{c \in \{ 0.1, 0.2 \cdots, 1.0 \} } ~~ \mathcal{L}(\theta, c) 
\end{align*}
where $\hat{f} = \calibrator(x_{1:k})$ is the predicted PC by the few-shot recalibrator, and $f$ is the groundtruth PC. This penalizes over-estimation more than under-estimation by setting the coefficient $\beta > 1.0$.

\subsection{Evaluation}
\label{ssec:eval}

Our few-shot recalibrator outputs a precision curve which is flexible and can be used to accomplish various downstream goals. We describe two of them here, along with the corresponding metrics that define success. 
We include another utility-based metric and its results in \cref{app:utility}. 
\paragraph{Achieving Target Precision}
For a given system, we may want to guarantee a minimum level of precision.
The goal, then, is to identify slice-specific confidence thresholds that achieve that level of precision without sacrificing much recall.
In this setting, we can directly use the predicted precision curve $\hat{g}$ as a lookup table and find the threshold that attains the target precision.
We evaluate performance by measuring the success rate of whether the selected threshold achieves the target precision on the ground-truth precision curve.

\paragraph{Reducing Calibration Error} 
Alternatively, the goal can be to reduce the system's calibration error.
For this setting, first we map the predicted precision curve $\hat{f}$ to the corresponding calibration curve, given the confidence scores of the predictions.
We do this as follows: let $\countt(a)$ denote the number of examples whose confidence exceeds $a$. For bin $B_i$, we have the upper $B_i.r$ and lower $B_i.l$ bounds on the confidence scores. We compute the accuracy for $B_i$:  $\acc(B_i) = \frac{\hat{f}(B_i.l) \countt(B_i.l) - \hat{f}(B_i.r) \countt(B_i.r)}{\countt(B_i.l) - \countt(B_i.r)}$, which along with the confidence $\conf(B_i)$, is sufficient to recover the calibration curve.
Once we have the calibration curve, we can apply histogram binning \citep{zadrozny2001calibrated} to map confidence scores to the corresponding accuracy, minimizing the calibration error.
We report ECE for this task.

\section{Experimental Setup} 
\subsection{Datasets} 
We evaluate our few-shot recalibrator on two datasets: MMLU \citep{hendrycks2021measuring} consists of multiple choice questions categorized into 57 different subjects (e.g. \emph{abstract algebra}, \emph{high school physics}, \emph{law}), each of which serves as a separate domain. XNLI \citep{conneau2018xnli} is a natural language inference task, where the model predicts if the given hypothesis entails, contradicts or is neutral to the corresponding premise. Examples are categorized into 10 genres (e.g. \emph{travel guides}, \emph{speeches}, etc.) in 15 languages each, for a total of 150 domains. 

We follow \cref{alg:1} to construct 20K slices for the training set and 2K unseen slices for the test set, ensuring that examples which appear in the test data's few-shot sets are held out from training. We also construct an \textsc{UNSEEN} test set for XNLI, where 10 domains are entirely held out from the training data and are used to construct a separate set of 2K mixtures.
For the main experiments we set $k=20$, and for ablation studies, we consider $k=\{5, 10, 20, 30\}$.

\begin{table*}[ht]
\centering
\small
\begin{tabular}{clcc|cc|cc|c}
\toprule
& \textbf{Target Precision} & \multicolumn{2}{c|}{\textbf{0.85}} & \multicolumn{2}{c|}{\textbf{0.9}} & \multicolumn{2}{c|}{\textbf{0.95}} &  \\
& & \textbf{Success} & \textbf{Recall} & \textbf{Success} & \textbf{Recall} & \textbf{Success} & \textbf{Recall} & \textbf{$L_2$} \\
\midrule
\multirow{5}{*}{\rotatebox{90}{\shortstack[c]{XNLI\\PaLM2-L}}}
& \textbf{\SampleAvg} & 0.47 & 0.86 & 0.55 & 0.71 & 0.62 & 0.42 & 0.001 \\
& \textbf{\DomainAvg} & 0.53 & 0.86 & 0.55 & 0.71 & 0.62 & 0.42 & 0.001 \\
& \textbf{Empirical} & 0.47 & 0.81 & 0.38 & 0.68 & 0.34 & 0.52 & 0.008 \\
& \textbf{\OursShort (Ours)} & \textbf{0.69} & 0.83 & \textbf{0.75} & 0.66 & \textbf{0.76} & 0.37 & \textbf{0.001} \\
& \OW{\textbf{Oracle}} & \OW{1.00} & \OW{0.85} & \OW{1.00} & \OW{0.7} & \OW{1.00} & \OW{0.45} & \OW{0.000} \\
\midrule
\multirow{5}{*}{\rotatebox{90}{\shortstack[c]{MMLU\\PaLM2-L}}}
& \textbf{\SampleAvg} & 0.64 & 0.95 & 0.64 & 0.88 & 0.60 & 0.75 & 0.006 \\
& \textbf{\DomainAvg} & 0.71 & 0.93 & 0.78 & 0.84 & 0.78 & 0.69 & 0.007 \\
& \textbf{Empirical} & 0.61 & 0.91 & 0.47 & 0.86 & 0.34 & 0.74 & 0.007 \\
& \textbf{\OursShort (Ours)} & \textbf{0.87} & 0.87 & \textbf{0.85} & 0.80 & \textbf{0.77} & 0.67 & \textbf{0.002} \\
& \OW{\textbf{Oracle}} & \OW{1.00} & \OW{0.91} & \OW{1.00} & \OW{0.85} & \OW{1.00} & \OW{0.74} & \OW{0.000}\\
\midrule
\multirow{5}{*}{\rotatebox{90}{\shortstack[c]{MMLU\\LLaMA-65B}}}
& \textbf{\SampleAvg} & 0.58 & 0.60 & 0.59 & 0.51 & 0.57 & 0.36 & 0.012 \\
& \textbf{\DomainAvg} & 0.72 & 0.57 & 0.80 & 0.41 & \textbf{0.99} & 0.02 & 0.012 \\
& \textbf{Empirical} & 0.43 & 0.58 & 0.40 & 0.48 & 0.34 & 0.40 & 0.023 \\
& \textbf{\OursShort (Ours)} & \textbf{0.90} & 0.50 & \textbf{0.89} & 0.39 & 0.80 & 0.23 & \textbf{0.006} \\
& \OW{\textbf{Oracle}} & \OW{1.00} & \OW{0.60} & \OW{1.00} & \OW{0.51} & \OW{1.00} & \OW{0.39} & \OW{0.000} \\
\bottomrule
\end{tabular}
\caption{Our few-shot recalibrator (FSC) has a higher success rate for identifying confidence thresholds that achieve a given target precision, as compared to the baselines, while maintaining reasonable recall. Here `Success' means the percentage of slices where we successfully achieve the target precision, and `L2' refers to distance between the predicted and gold precision curves. }
\label{tab:precision_success}
\vspace{-0.3cm}
\end{table*}

\begin{table*}[ht]
\centering
\small
\begin{tabular}{lccc|ccc|cccccccccc}
\toprule
& \multicolumn{3}{c}{XNLI (PaLM2-Large)} & \multicolumn{3}{|c}{MMLU (PaLM2-Large)} & \multicolumn{3}{|c}{MMLU (LLaMA-65B)}  \\
 & \textbf{ECE} & \textbf{Win\% } & \textbf{Lose\% } & \textbf{ECE} & \textbf{Win\%} & \textbf{Lose\%} & \textbf{ECE} & \textbf{Win\%} & \textbf{Lose\%} \\
\midrule
\textbf{Base} & 0.059 & 22 & 78 & 0.063 & 38 & 62 & 0.109 & 16 & 84 \\
\textbf{\SampleAvg} & 0.049 & 39 & 61 & 0.082 & 17 & 83 & 0.103 & 25 & 75 \\
\textbf{\DomainAvg} & 0.049 & 39 & 61 & 0.085 & 17 & 83 & 0.107 & 22 & 78 \\
\textbf{Empirical} & 0.094 & \phantom{0}9 & 91 & 0.078 & 29 & 71 & 0.122 & 14 & 86 \\

\textbf{TS (few-shot)} & 0.094 & \phantom{0}8 & 92 & 0.079 & 27 & 73 & 0.120 & 16 & 84 \\
\textbf{TS (all domains)} & 0.057 & 23 & 77 & 0.063 & 38 & 62 & 0.099 & 24 & 76 \\
\textbf{\OursShort (ours)} & \textbf{0.045} & \phantom{0}- & \phantom{0}- & \textbf{0.053} & \phantom{0}- & \phantom{0}- & \textbf{0.074} & \phantom{0}- & \phantom{0}- \\
\OW{\textbf{Oracle}} & \OW{0.011} &  \OW{99} & \OW{\phantom{0}1} & \OW{0.009} & \OW{100} & \OW{\phantom{0}0} & \OW{0.016} & \OW{100} & \OW{\phantom{0}0} \\
\bottomrule
\end{tabular}
\vspace{-0.2cm}
\caption{Our approach achieves the lowest calibration error (ECE), outperforming all baselines. 
Pairwise comparisons (with respect to our approach FSC) show that it has a lower ECE for most of the test slices, indicated by each baseline's lose percentage.
\textbf{Base} refers to the LM without any temperature scaling.}
\label{tab:ECE}
\vspace{-0.3cm}
\end{table*}

\subsection{Models} 
We train few-shot recalibrators for PaLM2-Large \citep{anil2023palm} and LLaMA-65B \citep{touvron2023llama} on MMLU and only PaLM2-Large, the best performing model, on XNLI. We also include recalibration results for LLaMA-30B in \cref{app:llama-30b}. 
Our recalibrator is a LLaMA-7B model, fine-tuned for $4$K steps for MMLU and $2$K for XNLI, both with a batch size of $16$, a learning rate of 2e-5 and a cosine learning rate schedule (see more details in \cref{app:hyperparameter}).
All finetuning experiments use $16$ A100-$40$GB GPUs. Recall from \cref{ssec:model}, our training objective is the asymmetric $L_2$ loss, and we set $\beta=5$ in all experiments.

\subsection{Baselines} 

We compare our few-shot recalibrator against the following baselines which output precision curves.

\textsc{Sample Average} is the precision curve of the combined distribution over all the domains based on the queries that appear in the training data. This baseline is not distribution-specific: it uses a single curve for all test set distributions.

\textsc{Domain Average} involves averaging the precision curves for each domain. 
Similar to sample averaging, this approach is not distribution-specific. 

\textsc{Empirical} uses the precision curve obtained from only the $k$ few-shot \emph{labeled} queries. Note that this baseline has an unfair advantage over other approaches, including ours, because it assumes access to the labels of the $k$ few-shot queries. 

\textsc{Oracle} is the ground-truth precision curve of the corresponding slice's distribution, and serves as a skyline for the best achievable performance for curve prediction approaches.

In the reducing calibration error setting, we compare our approach to the canonical recalibration method of temperature scaling \citep{guo2017calibration}. Temperature scaling (TS) uses a held out calibration set to select a temperature, and then applies that temperature to the test data. We compare against two variants of temperature scaling, and they differ in the choice of the calibration set. 

\textsc{TS (few-shot)} uses the $k$ few-shot examples with ground-truth labels as the calibration set. We run grid search on values for the temperature in $\{0.1, 0.2, \cdots, 1.9, 2.0, 3.0, 4.0, 5.0 \}$ to find one that minimizes ECE for the $k$ examples.

\textsc{TS (all domains)} uses the training data, combining all domains, as the calibration set.  Similarly, we run grid search on values for the temperature to minimize ECE for the entire training set. 

\begin{table*}[ht]
\centering
\small
\begin{tabular}{lcc|cc|cc|cc|c}
\toprule
\textbf{Target Precision} & \multicolumn{2}{c|}{\textbf{0.85}} & \multicolumn{2}{c|}{\textbf{0.9}} & \multicolumn{2}{c|}{\textbf{0.95}} &  \\
 & \textbf{Success} & \textbf{Recall} & \textbf{Success} & \textbf{Recall} & \textbf{Success} & \textbf{Recall} & \textbf{$L_2$} \\
\midrule
\textbf{$\SampleAvg$} & 0.60 & 0.86 & 0.63 & 0.70 & 0.38 & 0.42 & 0.002 \\
\textbf{$\DomainAvg$} & 0.65 & 0.85 & 0.63 & 0.70 & 0.38 & 0.42 & 0.002 \\
\textbf{Empirical} & 0.53 & 0.81 & 0.43 & 0.69 & 0.33 & 0.53 & 0.009 \\
\textbf{$\OursShort$(Ours)} & \textbf{0.79} & 0.83 & \textbf{0.74} & 0.67 & \textbf{0.69} & 0.34 & 0.001 \\
\OW{\textbf{Oracle}} & \OW{1.00} & \OW{0.87} & \OW{1.00} & \OW{0.72} & \OW{1.00} & \OW{0.43} & \OW{0.000} \\
\bottomrule
\end{tabular}
\vspace{-0.2cm}
\caption{Precision Success Rate On Unseen Domains from XNLI. Our approach achieves the highest success rate and lowest $L_2$ distance on previously unseen domains, without sacrificing much recall.}
\label{tab:unseen-precision}
\vspace{-0.3cm}
\end{table*}

\section{Main Results} 

\subsection{Achieving Target Precision} 

We first experiment with measuring the success rate of selecting a confidence threshold that achieves a given target precision on the slice's ground-truth precision curve.
As shown in \cref{tab:precision_success}, our few-shot recalibrator outperforms baselines by achieving a higher success rate for three different target precision values of $0.85$, $0.9$ and $0.95$.

\begin{figure*}
\vspace{-0.19cm}
\begin{minipage}[t]{0.3\linewidth}  %
    \centering
    \includegraphics[width=1.0\linewidth]{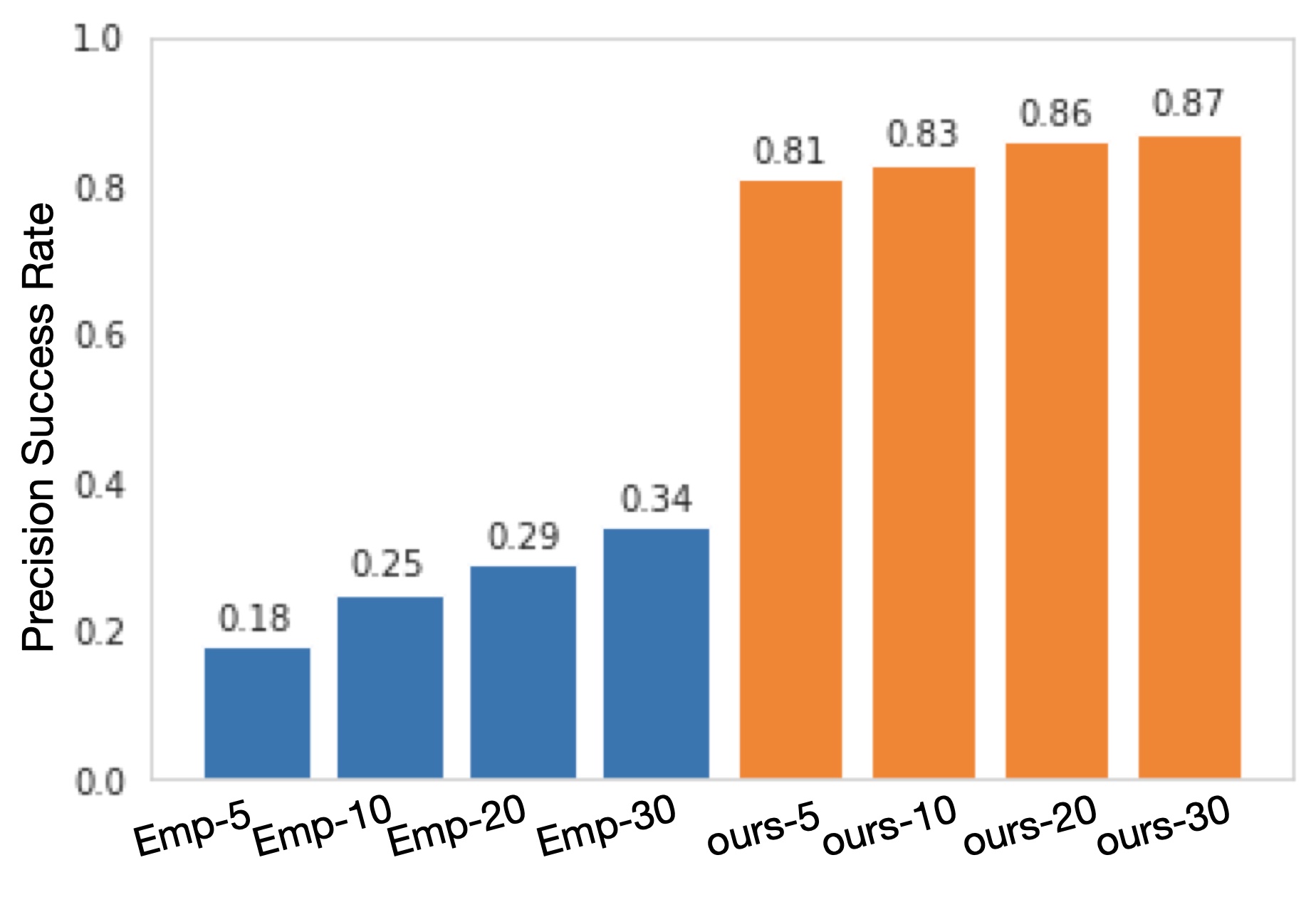}
    \caption{\label{fig:few-shotK} Our approach works well even with small few-shot sets.}
  \end{minipage}  \hfill
\begin{minipage}[t]{0.65\linewidth}
\centering
\includegraphics[width=1.0\linewidth]{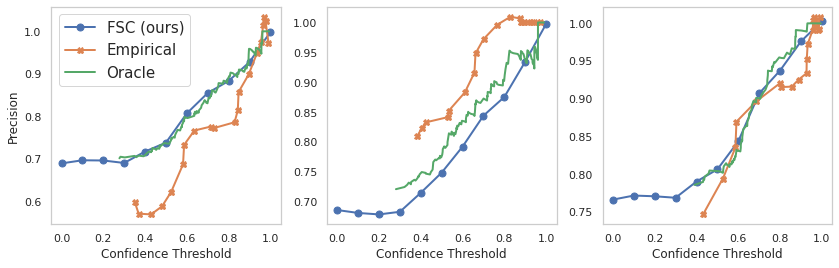}
\caption{\label{fig:qualitative} Examples of precision curves generated by the few-shot recalibrator, compared to the Empirical and Oracle curves. Our curves approximate the Oracle curves more closely.}
\end{minipage}
\vspace{-0.3cm}
\end{figure*}

In spite of the fact that the Empirical baseline has access to the few-shot example labels, our recalibrator consistently outperforms it by a large margin.
This shows that while the few-shot set itself is not sufficient for plotting a precision curve and selecting a slice-specific threshold, our recalibrator successfully learns to infer the full slice's distribution, and its corresponding precision curve, from this set.
This is also demonstrated in \cref{fig:qualitative}, where we show examples of precision curves generated by our few-shot recalibrator.
As we can see, the Empirical curve deviates far from the Oracle curve, while our recalibrator closely approximates it, and tends to upper bound it, as a consequence of our asymmetric training objective.

Our approach also outperforms the Sample and Domain averaging baselines in all settings but one: for a target precision of 0.95 when calibrating LLaMA-65B on MMLU. 
However, in this case Domain averaging achieves a high success rate of 0.99 by selecting an extremely high threshold and entirely sacrificing recall, down to 0.02. 
In contrast, our recalibrator strikes a better balance between achieving the target precision with a high success rate, while still maintaining reasonable recall.

\subsection{Reducing Calibration Error}

For the goal of reducing calibration error, we similarly find that our few-shot recalibrator outperforms baselines by achieving the lowest ECE score across various settings, as shown in \cref{tab:ECE}.
We also conduct a pairwise comparison and find that our recalibrator wins by achieving a lower ECE score most of the test slices as compared to all other approaches.

We find that the labeled few-shot set is not a useful proxy for the whole slice, since selecting a temperature based on this set for temperature scaling fails to improve ECE over the base LM with a temperature of 1.
We also find that selecting a single temperature for all slices, based on the broader distribution of the training set examples, is sub-optimal. 
In contrast, our few-shot recalibrator can provide slice-specific calibration which results in lower ECE.

\subsection{Extrapolation to Unseen Domains}
\label{ssec:extrapolation}

We also evaluate the extrapolation performance of our few-shot recalibrator. 
For this, we measure the success rate of achieving target precision on domains from XNLI that were \emph{unseen} in the training set. 
\cref{tab:unseen-precision} shows that our approach performs well on unseen domains as well, achieving the highest success rate of all curve prediction baselines, while maintaining a reasonable recall. 

\section{Ablation Studies} 
\begin{table*}[ht]
\centering
\small
\begin{tabular}{lccccccccccccc}
\toprule
 &  \multicolumn{2}{c}{\textbf{2 domains}} & \multicolumn{2}{c}{\textbf{3 domains}} & \multicolumn{2}{c}{\textbf{4 domains}} & \multicolumn{2}{c}{\textbf{5 domains}}  \\
& Success & Recall & Success & Recall & Success & Recall & Success & Recall &  \\
\midrule
\textbf{Empirical} & 0.39 & 0.68 & 0.40 & 0.65 & 0.34 & 0.71 & 0.29 & 0.70  \\
\textbf{$\OursShort$(ours)} & 0.76 & 0.66 & 0.75 & 0.65 & 0.77 & 0.65 & 0.71 & 0.66  \\
\OW{\textbf{Oracle}} & \OW{1} & \OW{0.70} & \OW{1} & \OW{0.69} & \OW{1} & \OW{0.71} & \OW{1} & \OW{0.70}  \\
\bottomrule
\end{tabular}
\caption{Model performance is robust to the number of domains included in the slice and the success rate does not vary significantly as the number of domains changes. }
\label{tab:mixtures}
\vspace{-0.35cm}
\end{table*}
We run all ablation experiments on the MMLU dataset, recalibrating the PaLM2-Large model.
\vspace{-0.2cm}
\paragraph{Number of few-shot examples} We examine the impact of the number of few-shot examples by experimenting with $k=\{5, 10, 20, 30\}$. As shown in \cref{fig:few-shotK}, the success rate of achieving target precision increases as we increase the number of few-shot examples for both the Empirical baseline and our few-shot recalibrator. Our approach with only 5 examples still achieves a high success rate of 0.81, suggesting it is highly suitable for settings with very small amounts of recalibration data.

\paragraph{Asymmetric vs. symmetric loss} The asymmetric objective penalizes over-estimation of precision more severely than under-estimation. In this ablation experiment, we verify the effectiveness of the asymmetric objective.  
We find that training our recalibrator with the asymmetric loss ($\beta=5$) results in a higher success rate of $0.85$ whereas the symmetric loss only achieves $0.68$, when aiming for a target precision of 90\%. 

\paragraph{Performance for different numbers of domains per slice} Our experiments involve constructing slices using different numbers of domains.
Here, we decompose target precision success rate results for mixtures containing 2, 3, 4 and 5 domains. \cref{tab:mixtures} shows that performance does not vary significantly across these settings.

\section{Related Work} 
Our approach draws inspiration from prior research on calibration, combining it with recent ideas from in-context few-shot learning \citep{wei2021finetuned,min2021metaicl}, especially \citet{lee2021neural}, which also trains a model over synthetically generated slices. Below, we further discuss relevant prior work on calibration for LMs and abstention.

\paragraph{Calibration for LMs} 
Recent work has found that pretrained language models appear mostly well-calibrated on broader distributions \citep{kadavath2022language, xiao2022uncertainty, kuhn2023semantic}, and can express their uncertainty in words \citep{lin2022teaching, mielke2022reducing,tian2023just,zhou2023navigating}. 
However, these models are still miscalibrated in some settings \citep{wang2020inference, stengeleskin2023calibrated}, and prior work has focused on recalibrating neural networks by temperature scaling \citep{guo2017calibration}, Platt scaling \citep{platt1999probabilistic}, isotonic regression \citep{niculescu2005predicting, zadrozny2002transforming}, or histogram binning \citep{kumar2019calibration, zadrozny2001calibrated}. 

However, the aforementioned works don't address  miscalibration within narrower slices, or slice-specific calibration. This problem has been more carefully studied outside LM research, such as for vision models \cite{Yu2022RobustCW} and from a theoretical angle \cite{pmlr-v80-hebert-johnson18a}. 

In this work, we address this problem for large language models. Also different from prior work is our ability to recalibrate using only \emph{unlabeled} few-shot examples, whereas previous methods have usually needed a non-trivial number of \emph{labeled} examples to achieve domain-specific calibration. \looseness=-1  

\vspace{-0.15cm}
\paragraph{Abstention}
When the model is not confident about a prediction, abstention or deferral to an expert are desirable alternatives compared to responding with the incorrect answer. In order to decide when to abstain, the line of work called rejection learning (or selective classification) focuses on \emph{jointly} learning a rejection function and a predictor \citep{tortorella2000binary,santos2005optimal,bartlett2008classification,Cortes2016Learning,geifman2017selective,fisch2022calibrated}. The rejection function decides when to abstain, and if the rejection function decides not to abstain, the predictor answers the question. 
In this paper, we freeze the base LM which functions as the predictor because it is computationally expensive to update a large model for downstream tasks. 
Instead, we make the abstention decision using our recalibrator and the raw confidence of the base LM. Specifically, we use the trained recalibrator to derive the confidence threshold above which the LM's prediction attains the target precision score. We also include experiments with a setup that closely matches the abstention setting in \cref{app:utility}.  
\vspace{-0.15cm}
\section{Conclusion and Future Work}

We have shown that while LMs appear to be well-calibrated on broad distributions, they remain miscalibrated for meaningful slices of that broader distribution.
To recalibrate them for each slice, we propose few-shot recalibration which takes few-shot, unlabeled queries and predicts a slice-specific precision curve. We then use the predicted precision curve for two downstream calibration tasks, finding that our approach consistently outperforms existing recalibration methods under all evaluation settings. 
Future work should study few-shot recalibration for natural language generation tasks, to steer model generated text to be more or less conservative, as well as apply this approach to a broader set of models, including instruction-tuned and RLHF models, and multimodal settings.

\section*{Limitation}
The problem setup here focuses on multiple-choice questions, for which there exists a unique correct answer and calibration is well-defined. However, one limitation of this paper is that we cannot handle open-ended responses, where there are exponential number of correct responses. We believe that calibrating open-ended responses remains a challenging yet important future research direction, and we include this idea in the future work section. 

\section*{Ethical Impact} 
Our paper focuses on adjusting the confidence of language models for each slice of distribution. One application is to define the slice based on demographics groups, and apply our approach to reduce calibration error for each demographics group. In this setting, our approach could improve fairness of the uncertainty calibration across different demographic groups. However, the proposed approach could also be misused by adversaries, if they adjust LM confidence in the direction that worsens calibration error for some targeted subgroups.

\section*{Acknowledgement} 
We thank Panupong Pasupat, Zhuyun Dai, Rishi Bommasani, Michael Xie, Tatsunori Hashimoto and Raphael Hoffmann for helpful comments and feedback. This work was done while Xiang Lisa Li interned at Google DeepMind. 

\bibliography{anthology,custom, iclr2024_conference, all}

\begin{thebibliography}{36}
\expandafter\ifx\csname natexlab\endcsname\relax\def\natexlab#1{#1}\fi

\bibitem[{Anil et~al.(2023)Anil, Dai, Firat, Johnson, Lepikhin, Passos,
  Shakeri, Taropa, Bailey, Chen, Chu, Clark, Shafey, Huang, Meier-Hellstern,
  Mishra, Moreira, Omernick, Robinson, Ruder, Tay, Xiao, Xu, Zhang, Abrego,
  Ahn, Austin, Barham, Botha, Bradbury, Brahma, Brooks, Catasta, Cheng, Cherry,
  Choquette-Choo, Chowdhery, Crepy, Dave, Dehghani, Dev, Devlin, Díaz, Du,
  Dyer, Feinberg, Feng, Fienber, Freitag, Garcia, Gehrmann, Gonzalez, Gur-Ari,
  Hand, Hashemi, Hou, Howland, Hu, Hui, Hurwitz, Isard, Ittycheriah, Jagielski,
  Jia, Kenealy, Krikun, Kudugunta, Lan, Lee, Lee, Li, Li, Li, Li, Li, Lim, Lin,
  Liu, Liu, Maggioni, Mahendru, Maynez, Misra, Moussalem, Nado, Nham, Ni,
  Nystrom, Parrish, Pellat, Polacek, Polozov, Pope, Qiao, Reif, Richter, Riley,
  Ros, Roy, Saeta, Samuel, Shelby, Slone, Smilkov, So, Sohn, Tokumine, Valter,
  Vasudevan, Vodrahalli, Wang, Wang, Wang, Wang, Wieting, Wu, Xu, Xu, Xue, Yin,
  Yu, Zhang, Zheng, Zheng, Zhou, Zhou, Petrov, and Wu}]{anil2023palm}
Rohan Anil, Andrew~M. Dai, Orhan Firat, Melvin Johnson, Dmitry Lepikhin,
  Alexandre Passos, Siamak Shakeri, Emanuel Taropa, Paige Bailey, Zhifeng Chen,
  Eric Chu, Jonathan~H. Clark, Laurent~El Shafey, Yanping Huang, Kathy
  Meier-Hellstern, Gaurav Mishra, Erica Moreira, Mark Omernick, Kevin Robinson,
  Sebastian Ruder, Yi~Tay, Kefan Xiao, Yuanzhong Xu, Yujing Zhang,
  Gustavo~Hernandez Abrego, Junwhan Ahn, Jacob Austin, Paul Barham, Jan Botha,
  James Bradbury, Siddhartha Brahma, Kevin Brooks, Michele Catasta, Yong Cheng,
  Colin Cherry, Christopher~A. Choquette-Choo, Aakanksha Chowdhery, Clément
  Crepy, Shachi Dave, Mostafa Dehghani, Sunipa Dev, Jacob Devlin, Mark Díaz,
  Nan Du, Ethan Dyer, Vlad Feinberg, Fangxiaoyu Feng, Vlad Fienber, Markus
  Freitag, Xavier Garcia, Sebastian Gehrmann, Lucas Gonzalez, Guy Gur-Ari,
  Steven Hand, Hadi Hashemi, Le~Hou, Joshua Howland, Andrea Hu, Jeffrey Hui,
  Jeremy Hurwitz, Michael Isard, Abe Ittycheriah, Matthew Jagielski, Wenhao
  Jia, Kathleen Kenealy, Maxim Krikun, Sneha Kudugunta, Chang Lan, Katherine
  Lee, Benjamin Lee, Eric Li, Music Li, Wei Li, YaGuang Li, Jian Li, Hyeontaek
  Lim, Hanzhao Lin, Zhongtao Liu, Frederick Liu, Marcello Maggioni, Aroma
  Mahendru, Joshua Maynez, Vedant Misra, Maysam Moussalem, Zachary Nado, John
  Nham, Eric Ni, Andrew Nystrom, Alicia Parrish, Marie Pellat, Martin Polacek,
  Alex Polozov, Reiner Pope, Siyuan Qiao, Emily Reif, Bryan Richter, Parker
  Riley, Alex~Castro Ros, Aurko Roy, Brennan Saeta, Rajkumar Samuel, Renee
  Shelby, Ambrose Slone, Daniel Smilkov, David~R. So, Daniel Sohn, Simon
  Tokumine, Dasha Valter, Vijay Vasudevan, Kiran Vodrahalli, Xuezhi Wang,
  Pidong Wang, Zirui Wang, Tao Wang, John Wieting, Yuhuai Wu, Kelvin Xu, Yunhan
  Xu, Linting Xue, Pengcheng Yin, Jiahui Yu, Qiao Zhang, Steven Zheng,
  Ce~Zheng, Weikang Zhou, Denny Zhou, Slav Petrov, and Yonghui Wu. 2023.
\newblock \href {http://arxiv.org/abs/2305.10403} {Palm 2 technical report}.

\bibitem[{Bartlett and Wegkamp(2008)}]{bartlett2008classification}
Peter~L Bartlett and Marten~H Wegkamp. 2008.
\newblock Classification with a reject option using a hinge loss.
\newblock \emph{Journal of Machine Learning Research (JMLR)}, 9(0):1823--1840.

\bibitem[{Conneau et~al.(2018)Conneau, Rinott, Lample, Williams, Bowman,
  Schwenk, and Stoyanov}]{conneau2018xnli}
Alexis Conneau, Ruty Rinott, Guillaume Lample, Adina Williams, Samuel Bowman,
  Holger Schwenk, and Veselin Stoyanov. 2018.
\newblock Xnli: Evaluating cross-lingual sentence representations.
\newblock In \emph{Empirical Methods in Natural Language Processing (EMNLP)},
  pages 2475--2485.

\bibitem[{Cortes et~al.(2016)Cortes, DeSalvo, and Mohri}]{Cortes2016Learning}
Corinna Cortes, Giulia DeSalvo, and Mehryar Mohri. 2016.
\newblock \href {https://api.semanticscholar.org/CorpusID:987180} {Learning
  with rejection}.
\newblock In \emph{International Conference on Algorithmic Learning Theory}.

\bibitem[{DeGroot and Fienberg(1983)}]{degroot1983forecasters}
Morris~H. DeGroot and Stephen~E. Fienberg. 1983.
\newblock The comparison and evaluation of forecasters.
\newblock \emph{Journal of the Royal Statistical Society. Series D (The
  Statistician)}, 32:12--22.

\bibitem[{Fisch et~al.(2022)Fisch, Jaakkola, and
  Barzilay}]{fisch2022calibrated}
Adam Fisch, Tommi~S. Jaakkola, and Regina Barzilay. 2022.
\newblock \href {https://openreview.net/forum?id=zFhNBs8GaV} {Calibrated
  selective classification}.
\newblock \emph{Transactions on Machine Learning Research}.

\bibitem[{Geifman and El-Yaniv(2017)}]{geifman2017selective}
Yonatan Geifman and Ran El-Yaniv. 2017.
\newblock Selective classification for deep neural networks.
\newblock In \emph{Advances in Neural Information Processing Systems
  (NeurIPS)}.

\bibitem[{Guo et~al.(2017)Guo, Pleiss, Sun, and
  Weinberger}]{guo2017calibration}
Chuan Guo, Geoff Pleiss, Yu~Sun, and Kilian~Q. Weinberger. 2017.
\newblock On calibration of modern neural networks.
\newblock In \emph{International Conference on Machine Learning (ICML)}, pages
  1321--1330.

\bibitem[{Gupta et~al.(2021)Gupta, Rahimi, Ajanthan, Mensink, Sminchisescu, and
  Hartley}]{gupta2021calibration}
Kartik Gupta, Amir Rahimi, Thalaiyasingam Ajanthan, Thomas Mensink, Cristian
  Sminchisescu, and Richard Hartley. 2021.
\newblock \href {https://openreview.net/forum?id=eQe8DEWNN2W} {Calibration of
  neural networks using splines}.
\newblock In \emph{International Conference on Learning Representations}.

\bibitem[{Hebert-Johnson et~al.(2018)Hebert-Johnson, Kim, Reingold, and
  Rothblum}]{pmlr-v80-hebert-johnson18a}
Ursula Hebert-Johnson, Michael Kim, Omer Reingold, and Guy Rothblum. 2018.
\newblock \href {https://proceedings.mlr.press/v80/hebert-johnson18a.html}
  {Multicalibration: Calibration for the ({C}omputationally-identifiable)
  masses}.
\newblock In \emph{Proceedings of the 35th International Conference on Machine
  Learning}, volume~80 of \emph{Proceedings of Machine Learning Research},
  pages 1939--1948. PMLR.

\bibitem[{Hendrycks et~al.(2021)Hendrycks, Burns, Basart, Zou, Mazeika, Song,
  and Steinhardt}]{hendrycks2021measuring}
Dan Hendrycks, Collin Burns, Steven Basart, Andy Zou, Mantas Mazeika, Dawn
  Song, and Jacob Steinhardt. 2021.
\newblock Measuring massive multitask language understanding.
\newblock In \emph{International Conference on Learning Representations
  (ICLR)}.

\bibitem[{Kadavath et~al.(2022)Kadavath, Conerly, Askell, Henighan, Drain,
  Perez, Schiefer, Hatfield-Dodds, DasSarma, Tran-Johnson, Johnston, El-Showk,
  Jones, Elhage, Hume, Chen, Bai, Bowman, Fort, Ganguli, Hernandez, Jacobson,
  Kernion, Kravec, Lovitt, Ndousse, Olsson, Ringer, Amodei, Brown, Clark,
  Joseph, Mann, McCandlish, Olah, and Kaplan}]{kadavath2022language}
Saurav Kadavath, Tom Conerly, Amanda Askell, Tom Henighan, Dawn Drain, Ethan
  Perez, Nicholas Schiefer, Zac Hatfield-Dodds, Nova DasSarma, Eli
  Tran-Johnson, Scott Johnston, Sheer El-Showk, Andy Jones, Nelson Elhage,
  Tristan Hume, Anna Chen, Yuntao Bai, Sam Bowman, Stanislav Fort, Deep
  Ganguli, Danny Hernandez, Josh Jacobson, Jackson Kernion, Shauna Kravec,
  Liane Lovitt, Kamal Ndousse, Catherine Olsson, Sam Ringer, Dario Amodei, Tom
  Brown, Jack Clark, Nicholas Joseph, Ben Mann, Sam McCandlish, Chris Olah, and
  Jared Kaplan. 2022.
\newblock \href {http://arxiv.org/abs/2207.05221} {Language models (mostly)
  know what they know}.

\bibitem[{Kuhn et~al.(2023)Kuhn, Gal, and Farquhar}]{kuhn2023semantic}
Lorenz Kuhn, Yarin Gal, and Sebastian Farquhar. 2023.
\newblock \href {https://openreview.net/forum?id=VD-AYtP0dve} {Semantic
  uncertainty: Linguistic invariances for uncertainty estimation in natural
  language generation}.
\newblock In \emph{The Eleventh International Conference on Learning
  Representations}.

\bibitem[{Kumar et~al.(2019)Kumar, Liang, and Ma}]{kumar2019calibration}
Ananya Kumar, Percy Liang, and Tengyu Ma. 2019.
\newblock Verified uncertainty calibration.
\newblock In \emph{Advances in Neural Information Processing Systems
  (NeurIPS)}.

\bibitem[{Lee et~al.(2021)Lee, Guu, He, Dozat, and Chung}]{lee2021neural}
Kenton Lee, Kelvin Guu, Luheng He, Tim Dozat, and Hyung~Won Chung. 2021.
\newblock Neural data augmentation via example extrapolation.
\newblock \emph{arXiv preprint arXiv:2102.01335}.

\bibitem[{Lin et~al.(2022)Lin, Hilton, and Evans}]{lin2022teaching}
Stephanie Lin, Jacob Hilton, and Owain Evans. 2022.
\newblock \href {https://openreview.net/forum?id=8s8K2UZGTZ} {Teaching models
  to express their uncertainty in words}.
\newblock \emph{Transactions on Machine Learning Research}.

\bibitem[{Mielke et~al.(2022)Mielke, Szlam, Dinan, and
  Boureau}]{mielke2022reducing}
Sabrina~J. Mielke, Arthur Szlam, Emily Dinan, and Y-Lan Boureau. 2022.
\newblock \href {https://doi.org/10.1162/tacl_a_00494} {{Reducing
  Conversational Agents’ Overconfidence Through Linguistic Calibration}}.
\newblock \emph{Transactions of the Association for Computational Linguistics},
  10:857--872.

\bibitem[{Min et~al.(2021)Min, Lewis, Zettlemoyer, and
  Hajishirzi}]{min2021metaicl}
Sewon Min, Mike Lewis, Luke Zettlemoyer, and Hannaneh Hajishirzi. 2021.
\newblock Metaicl: Learning to learn in context.
\newblock \emph{arXiv preprint arXiv:2110.15943}.

\bibitem[{Naeini et~al.(2015)Naeini, Cooper, and
  Hauskrecht}]{naeini2015obtaining}
Mahdi~Pakdaman Naeini, Gregory~F. Cooper, and Milos Hauskrecht. 2015.
\newblock Obtaining well calibrated probabilities using bayesian binning.
\newblock In \emph{Association for the Advancement of Artificial Intelligence
  (AAAI)}.

\bibitem[{Niculescu-Mizil and Caruana(2005)}]{niculescu2005predicting}
Alexandru Niculescu-Mizil and Rich Caruana. 2005.
\newblock Predicting good probabilities with supervised learning.
\newblock In \emph{Proceedings of the 22nd international conference on Machine
  learning}, pages 625--632.

\bibitem[{{OpenAI}(2023)}]{openai2023gpt4}
{OpenAI}. 2023.
\newblock {GPT}-4 technical report.
\newblock \emph{arXiv preprint arXiv:2303.08774}.

\bibitem[{Platt(1999)}]{platt1999probabilistic}
John Platt. 1999.
\newblock Probabilistic outputs for support vector machines and comparisons to
  regularized likelihood methods.
\newblock \emph{Advances in Large Margin Classifiers}, 10(3):61--74.

\bibitem[{Rasley et~al.(2020)Rasley, Rajbhandari, Ruwase, and
  He}]{rasley2020deepspeed}
Jeff Rasley, Samyam Rajbhandari, Olatunji Ruwase, and Yuxiong He. 2020.
\newblock \href {https://doi.org/10.1145/3394486.3406703} {Deepspeed: System
  optimizations enable training deep learning models with over 100 billion
  parameters}.
\newblock In \emph{Proceedings of the 26th ACM SIGKDD International Conference
  on Knowledge Discovery \& Data Mining}, KDD '20, page 3505–3506, New York,
  NY, USA. Association for Computing Machinery.

\bibitem[{Santos-Pereira and Pires(2005)}]{santos2005optimal}
Carla~M. Santos-Pereira and Ana~M. Pires. 2005.
\newblock \href {https://doi.org/https://doi.org/10.1016/j.patrec.2004.09.042}
  {On optimal reject rules and roc curves}.
\newblock \emph{Pattern Recognition Letters}, 26(7):943--952.

\bibitem[{Stengel-Eskin and Durme(2023)}]{stengeleskin2023calibrated}
Elias Stengel-Eskin and Benjamin~Van Durme. 2023.
\newblock \href {http://arxiv.org/abs/2211.07443} {Calibrated interpretation:
  Confidence estimation in semantic parsing}.

\bibitem[{Tian et~al.(2023)Tian, Mitchell, Zhou, Sharma, Rafailov, Yao, Finn,
  and Manning}]{tian2023just}
Katherine Tian, Eric Mitchell, Allan Zhou, Archit Sharma, Rafael Rafailov,
  Huaxiu Yao, Chelsea Finn, and Christopher~D. Manning. 2023.
\newblock \href {http://arxiv.org/abs/2305.14975} {Just ask for calibration:
  Strategies for eliciting calibrated confidence scores from language models
  fine-tuned with human feedback}.

\bibitem[{Tortorella(2000)}]{tortorella2000binary}
Francesco Tortorella. 2000.
\newblock An optimal reject rule for binary classifiers.
\newblock In \emph{Proceedings of the Joint IAPR International Workshops on
  Advances in Pattern Recognition}, page 611–620, Berlin, Heidelberg.
  Springer-Verlag.

\bibitem[{Touvron et~al.(2023)Touvron, Lavril, Izacard, Martinet, Lachaux,
  Lacroix, Rozière, Goyal, Hambro, Azhar, Rodriguez, Joulin, Grave, and
  Lample}]{touvron2023llama}
Hugo Touvron, Thibaut Lavril, Gautier Izacard, Xavier Martinet, Marie-Anne
  Lachaux, Timothée Lacroix, Baptiste Rozière, Naman Goyal, Eric Hambro,
  Faisal Azhar, Aurelien Rodriguez, Armand Joulin, Edouard Grave, and Guillaume
  Lample. 2023.
\newblock Llama: Open and efficient foundation language models.
\newblock \emph{arXiv}.

\bibitem[{Wang et~al.(2020)Wang, Tu, Shi, and Liu}]{wang2020inference}
Shuo Wang, Zhaopeng Tu, Shuming Shi, and Yang Liu. 2020.
\newblock \href {https://doi.org/10.18653/v1/2020.acl-main.278} {On the
  inference calibration of neural machine translation}.
\newblock In \emph{Proceedings of the 58th Annual Meeting of the Association
  for Computational Linguistics}, pages 3070--3079, Online. Association for
  Computational Linguistics.

\bibitem[{Wei et~al.(2021)Wei, Bosma, Zhao, Guu, Yu, Lester, Du, Dai, and
  Le}]{wei2021finetuned}
Jason Wei, Maarten Bosma, Vincent~Y. Zhao, Kelvin Guu, Adams~Wei Yu, Brian
  Lester, Nan Du, Andrew~M. Dai, and Quoc~V. Le. 2021.
\newblock Finetuned language models are zero-shot learners.
\newblock \emph{arXiv}.

\bibitem[{Wolf et~al.(2019)Wolf, Debut, Sanh, Chaumond, Delangue, Moi, Cistac,
  Rault, Louf, Funtowicz, and Brew}]{wolf2019huggingface}
Thomas Wolf, Lysandre Debut, Victor Sanh, Julien Chaumond, Clement Delangue,
  Anthony Moi, Pierric Cistac, Tim Rault, R{\'{e}}mi Louf, Morgan Funtowicz,
  and Jamie Brew. 2019.
\newblock \href {http://arxiv.org/abs/1910.03771} {Huggingface's transformers:
  State-of-the-art natural language processing}.
\newblock \emph{CoRR}, abs/1910.03771.

\bibitem[{Xiao et~al.(2022)Xiao, Liang, Bhatt, Neiswanger, Salakhutdinov, and
  Morency}]{xiao2022uncertainty}
Yuxin Xiao, Paul~Pu Liang, Umang Bhatt, Willie Neiswanger, Ruslan
  Salakhutdinov, and Louis-Philippe Morency. 2022.
\newblock \href {https://doi.org/10.18653/v1/2022.findings-emnlp.538}
  {Uncertainty quantification with pre-trained language models: A large-scale
  empirical analysis}.
\newblock In \emph{Findings of the Association for Computational Linguistics:
  EMNLP 2022}, pages 7273--7284, Abu Dhabi, United Arab Emirates. Association
  for Computational Linguistics.

\bibitem[{Yu et~al.(2022)Yu, Bates, Ma, and Jordan}]{Yu2022RobustCW}
Yaodong Yu, Stephen Bates, Yi-An Ma, and Michael~I. Jordan. 2022.
\newblock \href {https://api.semanticscholar.org/CorpusID:249394635} {Robust
  calibration with multi-domain temperature scaling}.
\newblock \emph{ArXiv}, abs/2206.02757.

\bibitem[{Zadrozny and Elkan(2001)}]{zadrozny2001calibrated}
Bianca Zadrozny and Charles Elkan. 2001.
\newblock Obtaining calibrated probability estimates from decision trees and
  naive bayesian classifiers.
\newblock In \emph{International Conference on Machine Learning (ICML)}, pages
  609--616.

\bibitem[{Zadrozny and Elkan(2002)}]{zadrozny2002transforming}
Bianca Zadrozny and Charles Elkan. 2002.
\newblock Transforming classifier scores into accurate multiclass probability
  estimates.
\newblock In \emph{International Conference on Knowledge Discovery and Data
  Mining (KDD)}, pages 694--699.

\bibitem[{Zhou et~al.(2023)Zhou, Jurafsky, and Hashimoto}]{zhou2023navigating}
Kaitlyn Zhou, Dan Jurafsky, and Tatsunori Hashimoto. 2023.
\newblock \href {http://arxiv.org/abs/2302.13439} {Navigating the grey area:
  Expressions of overconfidence and uncertainty in language models}.

\end{thebibliography}

\appendix
\newpage
\section{Hyperparameters}
\label{app:hyperparameter}
For inference of LLaMA-65B and LLaMA-30B to obtain the target precision curves, we use the deepspeed library \citep{rasley2020deepspeed} with 4 A-100 GPUs. For training the few-shot recalibrator, we finetune LLaMA-7B using the AdamW optimizer and a cosine learning rate schedule. We use a warmup ratio of 0.03, learning rate of $2e-5$, and batch size of 16. We train for 4K steps for the MMLU experiments and 2K steps for the XNLI experiments. Our fine-tuning is conducted on 16 A100 GPUs of 40GB memory, and we use Deepspeed Stage 3 to ensure the 7B model fits on GPU. Our implementation of inference and finetuning are based on the 
Hugging Face library \citep{wolf2019huggingface}.

\section{Maximizing Utility}  
\label{app:utility}
Recall in \cref{ssec:eval}, we studied two downstream evaluation tasks for calibration: (1) achieving target precision (2) reducing calibration error. Here, we discuss another downstream goal in practice: maximize the utility of a system, which consists of the abstention cost (sacrifices recall) and the error cost (sacrifices precision).
Inspired by the rejection learning framework \citep{Cortes2016Learning,bartlett2008classification}, we define a cost function that clearly specifies the trade-off:
incorrect predictions incur a cost of $1$ and abstaining incurs a cost $c \in [0, 1]$, while correct predictions incur no cost. For a fixed value for $c$, the goal is to maximize utility (i.e. negative cost).

Given the predicted precision curve $\precision_\theta$ and the raw confidence scores for predictions, let $\countt(t)$ denote the number of examples whose confidence exceeds $t$ and $N$ denote the total number of examples. Then, we estimate the cost at each threshold $t$ as $\text{Cost}(t) = (1-\precision_\theta(t)) \cdot \countt(t) + c \cdot (N-\countt(t))$, where the first term accounts for incorrect predictions and the second term accounts for abstentions.
And we find the optimal threshold $t^*$ that minimizes $\text{Cost}(t)$ via a grid search over $t \in [0, 1]$. To evaluate the goodness of the selected threshold $t^*$, we assume access to labeled data, and measure the empirical utility achieved by abstaining when the model's confidence is lower than the selected threshold and making a prediction otherwise.

\label{ssec:utility_results}
\begin{table*}[ht]
\centering
\small
\begin{tabular}{lcc|cc|cc}
\toprule
 & \multicolumn{2}{c|}{XNLI (PaLM2-Large)} & \multicolumn{2}{c|}{MMLU (PaLM2-Large)} & \multicolumn{2}{c}{MMLU (LLaMA-65B)} \\
& $c=0.4$ & $c=0.6$ & $c=0.4$ & $c=0.6$ & $c=0.4$ & $c=0.6$ \\
\midrule
\textbf{Abstain} & -0.224 & -0.240 & -0.162 & \textbf{-0.188} & -0.315 & -0.390 \\
\textbf{\SampleAvg} & -0.206 & -0.219 & -0.169 & -0.197 & -0.289 & -0.382 \\
\textbf{\DomainAvg} & -0.206 & -0.219 & -0.171 & -0.197 & -0.289 & -0.388 \\
\textbf{Empirical} & -0.208 & -0.225 & -0.164 & -0.190 & -0.293 & \textbf{-0.372} \\
\textbf{\OursShort (Ours)} & \textbf{-0.202} & \textbf{-0.218} & \textbf{-0.157} & -0.189 & \textbf{-0.284} & \textbf{-0.372} \\
\OW{\textbf{Oracle}} & \OW{-0.192} & \OW{-0.213} & \OW{-0.150} & \OW{-0.180} & \OW{-0.277} & \OW{-0.358} \\
\bottomrule
\end{tabular}
\caption{ Our few-shot recalibrator is better at maximizing utility, and thus, finding the right balance between abstaining and making predictions.
}
\label{tab:utility}
\vspace{-0.4cm}
\end{table*}

\begin{table*}[ht]
\centering
\begin{tabular}{clcccc|cccc}
\toprule
& & \multicolumn{4}{c}{$c=0.4$} & \multicolumn{4}{|c}{$c=0.6$} \\
& & \textbf{Utility} & \textbf{Win} & \textbf{Tie} & \textbf{Lose} & \textbf{Utility} & \textbf{Win} & \textbf{Tie} & \textbf{Lose} \\
\midrule
\multirow{6}{*}{\rotatebox{90}{\shortstack[c]{XNLI\\PaLM2-L}}} & 
\textbf{Abstain} & -0.224 & 0.4 & 0.0005 & 0.5995 & -0.24 & 0.398 & 0.0035 & 0.5985 \\
& \textbf{Curve agg} & -0.206 & 0.183 & 0.3795 & 0.4375 & -0.219 & 0.218 & 0.4975 & 0.2845 \\
& \textbf{few-shot} & -0.208 & 0.332 & 0.0775 & 0.5905 & -0.225 & 0.299 & 0.246 & 0.455 \\
& \textbf{\OursShort (Ours)} & -0.202 & 0 & 1 & 0 & -0.218 & 0 & 1 & 0 \\
& \textbf{Oracle} & -0.192 & 0.851 & 0.098 & 0.051 & -0.213 & 0.709 & 0.22 & 0.071 \\
\midrule
\multirow{6}{*}{\rotatebox{90}{\shortstack[c]{MMLU\\PaLM2-L}}}
& \textbf{Abstain} & -0.162 & 0.484 & 0.0015 & 0.5145 & -0.188 & 0.5085 & 0.0015 & 0.49 \\
& \textbf{Curve\_agg} & -0.171 & 0.188 & 0.2005 & 0.6115 & -0.197 & 0.176 & 0.2355 & 0.5885 \\
& \textbf{few-shot} & -0.164 & 0.3095 & 0.0885 & 0.602 & -0.19 & 0.4205 & 0.0885 & 0.491 \\
& \textbf{\OursShort (Ours)} & -0.157 & 0 & 1 & 0 & -0.189 & 0 & 1 & 0 \\
& \textbf{Oracle} & -0.15 & 0.862 & 0.096 & 0.042 & -0.18 & 0.823 & 0.124 & 0.053 \\
\midrule
\multirow{6}{*}{\rotatebox{90}{\shortstack[c]{MMLU\\LLaMA-65B}}}
& \textbf{Abstain} & -0.315 & 0.322 & 0.001 & 0.677 & -0.39 & 0.401 & 0.002 & 0.597 \\
& \textbf{Curve\_agg} & -0.289 & 0.2715 & 0.2135 & 0.515 & -0.388 & 0.225 & 0.1245 & 0.6505 \\
& \textbf{few-shot} & -0.293 & 0.3105 & 0.091 & 0.5985 & -0.372 & 0.448 & 0.1305 & 0.4215 \\
& \textbf{\OursShort (Ours)} & -0.284 & 0 & 1 & 0 & -0.372 & 0 & 1 & 0 \\
& \textbf{Oracle} & -0.277 & \OW{0.787} & 0.139 & 0.074 & -0.358 & 0.817 & 0.088 & 0.095 \\
\bottomrule
\end{tabular}
\caption{Additional utility results, including the pairwise comparisons win/tie/lose rate compared to our approach. Overall, our few-shot recalibrator outperforms all baselines in achieving the highest utility scores, and more winning percentages.}
\label{app_tab:utility}
\end{table*}

For the utility maximization setting, we experiment with two values of the abstention costs, $c=0.4$ which favors abstaining more (i.e. precision) and $c=0.6$ which favors answering more (i.e. recall). These two settings evaluate each method's flexibility to balance different trade-offs between precision and recall.
As shown in \cref{tab:utility}, we find that our few-shot calibrator strikes a good trade-off between precision and recall for both settings, consistently achieving a higher utility as compared to baselines, including the Abstain model.

Additionally, we provide pairwise comparison results that contains win/tie/lose rate of each baseline v.s. our approach in \cref{app_tab:utility}.

\section{Additional Results (LLaMA-30B)}
\label{app:llama-30b}
In addition to LLaMA-65B and PaLM2-Large, we also apply our few-shot recalibrator approach to LLaMA-30B to study the impact of model scales. See results in \cref{app:tab:30b_success}, \cref{app:tab:30b_ece}, and \cref{app:tab:30b_utility}. Compared to other base models (LLaMA-65B model and PaLM2-Large), we observe similar trends in the minimizing ECE and maximizing utility experiment: We find that our approach outperform all baselines in achieving the lowest calibration error with the highest win rate (\cref{app:tab:30b_ece}). In addition, our approach outperform all baselines in selecting an abstention threshold that yields the highest utility score (\cref{app:tab:30b_utility}). The only exception happens for the precision success rate experiment. Unlike the results of LLaMA-65B where our few-shot recalibrator outperform all the baselines including $\DomainAvg$, for LLaMA-30B, $\DomainAvg$ achieves higher success rate than our few-shot recalibrator. The gap is particularly large for a target precision of 0.95. We hypothesis that this is because the LLaMA-30B suffers from lower accuracy compared to larger models. Thus, in the training data, the groundtruth precision curve of many custom distributions fail to hit the 95\% precision level, leading to a sparsity of training data that hits the 95\% precision level. As a result, when we try to infer about 95\% precision level at inference time, the model predictions are more prone to error. 
\begin{table*}[htbp]
    \centering
    \begin{tabular}{llcc|cc|cc|c}
        \toprule
        & \textbf{Target Precision} & \multicolumn{2}{c|}{\textbf{0.85}} & \multicolumn{2}{c|}{\textbf{0.9}} & \multicolumn{2}{c|}{\textbf{0.95}} &  \\
& & \textbf{Success} & \textbf{Recall} & \textbf{Success} & \textbf{Recall} & \textbf{Success} & \textbf{Recall} & \textbf{$L_2$} \\
\midrule
\multirow{5}{*}{\rotatebox{90}{\shortstack[c]{MMLU\\LLaMA-30B}}}
& \textbf{\SampleAvg} & 0.57 & 0.45 & 0.58 & 0.36 & 0.59 & 0.26 & 0.012 \\
& \textbf{\DomainAvg} & 0.76 & 0.38 & 0.72 & 0.32 & 0.94 & 0.09 & 0.013 \\
& \textbf{Empirical} & 0.36 & 0.5 & 0.34 & 0.42 & 0.28 & 0.35 & 0.030 \\
& \textbf{FSC (ours)} & 0.75 & 0.35 & 0.68 & 0.26 & 0.52 & 0.16 & 0.007 \\
& \textbf{Oracle} & 1 & 0.46 & 1 & 0.38 & 1 & 0.28 & 0 \\
        \bottomrule
    \end{tabular}
    \caption{\label{app:tab:30b_success} Precision Success Rate for LLaMA-30B on MMLU.  $\DomainAvg$ achieves higher success rate than our few-shot recalibrator. The gap is particularly large for a target precision of 0.95. We hypothesizes that this is because the LLaMA-30B suffers from lower accuracy compared to larger models (LLaMA-65B). Thus, in the training data, the groundtruth precision curve of many custom distributions fail to hit the 95\% precision level, leading to a sparsity of training data that hits the 95\% precision level. As a result, when we try to infer about 95\% precision level at inference time, the model predictions are more prone to error.} 
\end{table*}

\begin{table*}[htbp]
    \centering
    \begin{tabular}{lccc}
        \toprule
        Method & ECE & win\% & lose\% \\
        \midrule
        \textbf{Base} & 0.093 & 0.2425 & 0.7575 \\
        \textbf{\SampleAvg} & 0.106 & 0.2325 & 0.7675 \\
        \textbf{\DomainAvg} & 0.109 & 0.192 & 0.808 \\
        \textbf{Empirical} & 0.131 & 0.091 & 0.909 \\
        \textbf{TS (few-shot)} & 0.117 & 0.187 & 0.813 \\
        \textbf{TS (all domains)} & 0.090 & 0.283 & 0.717 \\
        \textbf{FSC(ours)} & 0.074 & - & - \\
        \textbf{Oracle} & 0.016 & 0.9975 & 0.0025 \\
        \bottomrule
    \end{tabular}
\caption{\label{app:tab:30b_ece} ECE for LLaMA-30B on MMLU. Our approach outperforms all the baselines in achieving the lowest calibration error with the highest win rate.} 
\end{table*}

\begin{table*}[htbp]
    \centering
    
    \begin{tabular}{llcccc|cccc}
        \toprule
        & & \multicolumn{4}{c}{$c=0.4$} & \multicolumn{4}{|c}{$c=0.6$} \\
& & \textbf{Utility} & \textbf{Win} & \textbf{Tie} & \textbf{Lose} & \textbf{Utility} & \textbf{Win} & \textbf{Tie} & \textbf{Lose} \\
\midrule
\multirow{6}{*}{\rotatebox{90}{\shortstack[c]{XNLI\\PaLM2-L}}} & \textbf{Abstain} & -0.352 & 0.3065 & 0.001 & 0.6925 & -0.437 & 0.4595 & 0.002 & 0.5385 \\
& \textbf{\SampleAvg} & -0.326 & 0.231 & 0.212 & 0.557 & -0.443 & 0.2445 & 0.1345 & 0.621 \\
& \textbf{\DomainAvg} & -0.329 & 0.185 & 0.145 & 0.67 & -0.451 & 0.1985 & 0.0905 & 0.711 \\
& \textbf{Empirical} & -0.329 & 0.279 & 0.0805 & 0.6405 & -0.431 & 0.4105 & 0.1065 & 0.483 \\
& \textbf{FSC(ours)} & -0.319 & 0 & 1 & 0 & -0.428 & 0 & 1 & 0 \\
& \textbf{Oracle} & -0.311 & 0.8125 & 0.13 & 0.0575 & -0.416 & 0.8215 & 0.099 & 0.0795 \\
        \bottomrule
    \end{tabular}
\caption{\label{app:tab:30b_utility} Utility Scores for LLaMA-30B on MMLU. Our approach outperforms all baselines in selecting abstention thresholds that yield the highest utility scores.} 
\end{table*}

\section{Additional Results (Extrapolation)}
Recall in \cref{ssec:extrapolation}, we show our few-shot recalibrator extrapolates well to unseen domains as demonstrated by the precision success rate experiments. Here, we provide more evidence, demonstrated by the ECE results in \cref{app_tab:unseen_ece}. Same as the trend in the precision experiment, our approach outperforms all the baselines in achieving the lowest calibration error and more winning percentages in pairwise comparison. 
\begin{table*}[ht]
\centering
\begin{tabular}{lccc}
\toprule
\textbf{Method} & \textbf{ECE} & \textbf{Win} & \textbf{Lose} \\
\midrule
\textbf{Base} & 0.064 & 0.268 & 0.732 \\
\textbf{\SampleAvg} & 0.052 & 0.4525 & 0.5475 \\
\textbf{\DomainAvg} & 0.052 & 0.444 & 0.556 \\
\textbf{Empirical} & 0.093 & 0.115 & 0.885 \\
\textbf{TS (few-shot)} & 0.095 & 0.1285 & 0.8715 \\
\textbf{TS (all domains)} & 0.061 & 0.3155 & 0.6845 \\
\textbf{FSC (ours)} & 0.049 & - & - \\
\textbf{Oracle} & 0.011 & 0.9965 & 0.0035 \\
\bottomrule
\end{tabular}
\caption{Unseen ECE Evaluation. Our approach outperforms all the baselines in achieving the lowest calibration error and more winning percentages in pairwise comparison. }
\label{app_tab:unseen_ece}
\end{table*}

\section{Licenses for Scientific Artifacts}
All artifacts are used as intended, evaluating using the datasets, and performing training/inference using the models. 
\begin{itemize}
    \item MMLU dataset (MIT License)
    \item XNLI dataset (Creative Commons Public)
    \item LLaMA models (LLAMA 2 COMMUNITY LICENSE AGREEMENT) 
\end{itemize}

\end{document}